\documentclass[sigconf]{acmart}
\AtBeginDocument{%
  \providecommand\BibTeX{{%
    \normalfont B\kern-0.5em{\scshape i\kern-0.25em b}\kern-0.8em\TeX}}}

\setcopyright{acmlicensed}
\copyrightyear{2024}
\acmYear{2024}
\acmDOI{XXXXXXX.XXXXXXX}

\acmConference[ICMI'24]{26th ACM International Conference on Multimodal Interaction}{Nov 04--08,
  2024}{San Jos\'e, Costa Rica}
\acmISBN{978-1-4503-XXXX-X/18/06}

\usepackage{graphicx}
\usepackage{subcaption}

\begin{document}

\title{3D Gaze Tracking for Studying Collaborative Interactions in Mixed-Reality Environments}

\author{Eduardo Davalos}
\email{eduardo.davalos.anaya@vanderbilt.edu}
\orcid{0000-0001-7190-7273}
\affiliation{%
  \institution{Vanderbilt University}
  \city{Nashville}
  \state{TN}
  \country{USA}
}

\author{Yike Zhang}
\email{yike.zhang@vanderbilt.edu}
\orcid{0000-0003-3503-2996}
\affiliation{%
  \institution{Vanderbilt University}
  \city{Nashville}
  \state{TN}
  \country{USA}
}

\author{Ashwin T. S.}
\email{ashwindixit9@gmail.com}
\orcid{0000-0002-1690-1626}
\affiliation{%
  \institution{Vanderbilt University}
  \city{Nashville}
  \state{TN}
  \country{USA}
}

\author{Joyce H. Fonteles}
\email{joyce.h.fonteles@vanderbilt.edu}
\orcid{0000-0001-9862-8960}
\affiliation{%
  \institution{Vanderbilt University}
  \city{Nashville}
  \state{TN}
  \country{USA}
}

\author{Umesh Timalsina}
\email{umesh.timalsina@vanderbilt.edu}
\orcid{0000-0002-5430-3993}
\affiliation{%
  \institution{Vanderbilt University}
  \city{Nashville}
  \state{TN}
  \country{USA}
}

\author{Gautam Biswas}
\email{gautam.biswas@vanderbilt.edu}
\orcid{0000-0002-2752-3878}
\affiliation{
  \institution{Vanderbilt University}
  \city{Nashville}
  \state{TN}
  \country{USA}
}

\renewcommand{\shortauthors}{Davalos et al.}

\begin{abstract}

This study presents a novel framework for 3D gaze tracking tailored for mixed-reality settings, aimed at enhancing joint attention and collaborative efforts in team-based scenarios. Conventional gaze tracking, often limited by monocular cameras and traditional eye-tracking apparatus, struggles with simultaneous data synchronization and analysis from multiple participants in group contexts. Our proposed framework leverages state-of-the-art computer vision and machine learning techniques to overcome these obstacles, enabling precise 3D gaze estimation without dependence on specialized hardware or complex data fusion. Utilizing facial recognition and deep learning, the framework achieves real-time, tracking of gaze patterns across several individuals, addressing common depth estimation errors, and ensuring spatial and identity consistency within the dataset. Empirical results demonstrate the accuracy and reliability of our method in group environments. This provides mechanisms for significant advances in behavior and interaction analysis in educational and professional training applications in dynamic and unstructured environments.


\end{abstract}

\ccsdesc[500]{Applied computing~Collaborative learning}
\ccsdesc[300]{Computer systems organization~Real-time system architecture}
\ccsdesc[500]{Human-centered computing~Collaborative interaction}

\keywords{eye-tracking, gaze, collaborative, joint attention, automated}

\begin{teaserfigure}
  \includegraphics[width=\textwidth]{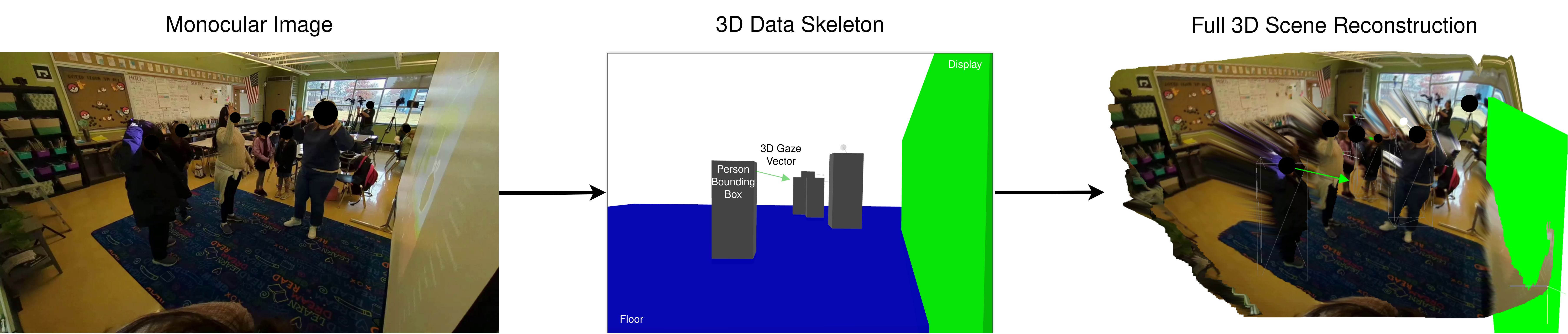}
  \caption{3 sequential representations of a scene: monocular, 3D skeleton data, and full 3D reconstruction to perform gaze analysis.}
  \Description[3 items from left to right form 2D, 3D skeleton, and full 3D mesh]{}
  \label{fig:teaser}
\end{teaserfigure}

\received{10 May 2024}
\received[revised]{1 June 2024}
\received[accepted]{1 July 2024}

\maketitle


\section{Introduction}


Eye-tracking technology has significantly advanced our understanding of how humans perceive and process information, particularly in situations where cognitive processing and information acquisition are essential components for completing tasks. This technology is widely used in educational and professional training environments, offering insights into the intricate relationship between cognition and task performance \cite{rosch2013review}.

In educational environments, eye-tracking technology informs researchers and educators about students' visual attention and engagement patterns associated with successful learning and comprehension. Therefore, insights gained from eye-tracking studies contribute to developing educational tools and methodologies that are more engaging and effective in improving student learning outcomes. Additionally, a significant body of eye-tracking research examines the interplay between individual and collective gaze behaviors within collaborative problem-solving settings. Comprehending these interactions is instrumental in enhancing team-based strategies and communication techniques, as eye-tracking data elucidates the distribution of visual attention among team members and their reactions to shared visual signals during cooperative activities \cite{schneider2018leveraging}.


A lot of current eye-tracking research is predominantly confined to individual settings \cite{Deng2023ALearning}. Collaborative and team-based studies present many challenges, such as the alignment of data across multiple devices, limited support from eye-tracking manufacturers, and the substantial costs associated with engineering and developing specialized hardware and software to support data collection and analyses. Most eye-tracking devices are tailored for single-user applications and do not easily generalize to studying multi-user interactions. Additionally, the development of new methods and technologies involves significant engineering expenses in terms of human hours and technical expertise. This issue is compounded by the traditionally poor scalability and high costs of eye-tracking equipment.


This research addresses the complexities of collaborative gaze analysis for children enacting scientific processes in mixed-reality environments \cite{fontelesAIED2024}. Our framework enhances the conventional gaze-tracking approaches by incorporating 3D objects-of-interest (OOI) encoding. This extension progresses from the traditional 2D areas-of-interest (AOI) to a more dynamic 3D space, offering detailed insights into both the social and environmental context of gaze behavior. The key contributions of our work are:

\begin{itemize}
    \item \textbf{Facial Recognition} We implemented face recognition to consistently track individuals as they move around in physical space, even when they occlude one another, thus enabling continuous monitoring of gaze patterns for multiple users.
    \item \textbf{3D Reconstruction and Gaze Reprojection} Our approach includes a sophisticated 3D reconstruction of the environment, allowing us to reproject gaze data accurately within this space. This method provides a realistic depiction of where individuals are looking, enhancing the accuracy of gaze tracking.
    \item \textbf{Social Network Analysis} We use the continuous and simultaneous gaze tracking of multiple individuals to perform social network analysis and visualize the social dynamics among the participating students.
\end{itemize}



Our framework enhances 3D reconstruction and robustly encodes OOIs in a scene, capturing dynamic user movements and static objects like screens and furniture for in-depth identity-aware gaze tracking, which is essential for deciphering social interactions and engagement by monitoring how participants view each other and interact with room objects \cite{danish2020learning}. These capabilities ensure precise identification and analysis of gaze targets in complex, noisy environments, such as classrooms. A detailed timeline of OOI gaze events in real-world scenarios demonstrates the framework's practicality. For transparency and to promote further development, we have made the entire codebase for this project open-source, available in <<Github>> repository.

\section{Background}

The field of eye-tracking and gaze estimation has witnessed significant advances, evolving from hardware-centric approaches, like the ones developed by Tobii and Eyeware, to sophisticated computer vision techniques. The emergence of datasets like GazeFollow \cite{nips15_recasens} has led to the development of algorithms capable of predicting gaze within complex scenes. Technologies like MIDAS \cite{Ranftl2022TowardsTransfer} and transformer-based approaches \cite{tu2022end} have further refined accuracy by integrating depth and object recognition, although challenges like depth noise and computational intensity remain. Pre-processing techniques have been instrumental in translating gaze data into behavioral insights, particularly within social network analysis, thereby enhancing our understanding of interaction dynamics across both digital and physical domains.

\subsection{Eye-Tracking \& Gaze Estimation}

Eye-tracking, an elaborate 3D problem, has traditionally been addressed by eye-tracking manufacturers, such as Tobii, EyeWare, and EyeLink. They include on-screen bars, webcams, and glasses. These plug-and-play solutions have enriched the field of eye-tracking research, opened many new directions, and accelerated eye-tracking research. Within the educational domain, these tools and equipment have been successfully employed to use gaze tracking analysis to understand individuals' learning behaviors \cite{Rajendran2018_Eye-gaze_Learning_Gains,tai2006exploration, festor2022midas}.


Eye-tracking devices may be limited in their applications, especially when the environment, such as a collaborative learning setting, has requirements beyond their initial design specifications. Although some studies have attempted to use these devices in group environments \cite{Jing2021EyeCollaboration, Lamsa2022TheTracking, Villamor2019GazeAnalysis}, they have not produced the same depth of analysis as individual-focused research. This shortfall is largely due to the complexities in integrating and synchronizing gaze data from multiple sensors. Such integration typically requires sophisticated engineering solutions that existing eye-tracking and multimodal software, such as iMotions\texttrademark\footnote{\url{https://imotions.com/}}, do not support. Consequently, conducting collaborative gaze studies often incurs significant development costs in hardware and software, which reduces the robustness and scalability of these approaches.

Recent developments in computer vision have produced innovative systems designed to address the challenges of 3D gaze estimation. Notable examples, Gaze360 \cite{Kellnhofer2019Gaze360:Wild}, L2CS-Net \cite{Abdelrahman2023L2CS-NetOfficial}, and MCGaze \cite{Guan2023End-to-EndContext}, represent significant advances in developing robust and scalable approaches to gaze tracking and analysis. These systems harness deep learning algorithms to estimate gaze vectors' pitch and yaw from standard monocular camera feeds. Such advances negate the need for specialized eye-tracking hardware and the need for complex data synchronization algorithms, facilitating the tracking of multiple users' gazes concurrently and making it easier to run education studies in classrooms.

The field of eye-tracking has recently expanded to include innovative research in \textit{gaze target detection}, an approach aimed at discerning the focal point of a person's gaze within a given scene. This technique involves estimating the gaze-fixed object and the gaze vector using deep learning (DL) methods. The GazeFollow dataset \cite{nips15_recasens} has been instrumental in propelling this research area forward by providing a benchmark dataset for the training and evaluation of DL models tailored to this task. For the gaze target detection task, a single monocular image serves as the input. This image is analyzed to predict the exact point within the visual field where the gaze is directed. A pioneering approach by \citet{Tonini2023Object-aware} has furthered this technique by incorporating MIDAS relative depth estimation models. These models enhance the input monocular images with 3D scene information, creating a more comprehensive understanding of the space within the image.

Following the depth estimation, a series of transformer modules are applied to perform a sequence of tasks. Initially, these transformers identify and classify objects within the image. Subsequently, they generate field-of-vision (FOV) cones, which help in visualizing the possible areas within the scene that might be the focus of the gaze. Finally, the transformers link the gaze vector to these objects, effectively predicting the gaze target. However, this approach does come with trade-offs. Many of the gaze target estimation methods are not geared for temporal analysis; therefore, their output is not temporally consistent or congruent. While it bypasses the need for more complex camera setups to achieve a 3D understanding of the scene, it may suffer from high-depth noise. This noise can lead to inaccuracies in scenarios where precise depth information is crucial. Moreover, the computational expense of depth estimation methods needed for accurate 3D scene reconstruction can be substantial, potentially limiting the method's applicability in real-time or on-device applications. Despite these challenges, the advances in gaze target detection exemplify the dynamic nature of eye-tracking research and its potential to enhance our interaction with and understanding of complex visual environments.

\subsection{Pre-processing Gaze}

The literature on eye-tracking features various processing techniques that correlate gaze data with learning outcomes \cite{Rajendran2018_Eye-gaze_Learning_Gains}, cognitive workload estimation \cite{Ktistakis2022COLET:Eye-tracking}, and behaviors \cite{srivastava2023characteristics}. Eye-tracking devices, such as the EyeLink 2000, often produce data at rates exceeding 150 Hz, leading to the development of numerous preprocessing techniques. These techniques refine the data into more manageable representations and metrics that are easier to analyze and link to educational or training contexts. According to Srivastava et al. \cite{srivastava2018}, eye-tracking features are categorized into three levels: LOW, which includes fixation and saccade metrics; MID, encompassing gaze radial direction and patterns; and HIGH, which involves encoding areas-of-interest (AOIs). These categories are utilized for various applications, including training machine learning algorithms, computing statistics, and generating visualizations.

Each category of eye-tracking features offers specific advantages, limitations, and demands for computational resources. LOW and MID level features do not require additional data and are less computationally intensive to process. Conversely, HIGH-level features, which connect gaze data directly with relevant visual elements within the environment, provide deeper insights but are more complex to compute. AOI encoding, for example, involves defining the geometry of an AOI -- typically a rectangle or a circle -- and determining whether a gaze point falls within it \cite{davalos2023identifying}. AOIs are further classified as dynamic or static, with dynamic AOIs presenting significant challenges as they can move within the tracking session. Depending on the nature of the AOI -- such as a person or an object like a television -- and the specific eye-tracking system employed, tracking dynamic AOIs may require advanced deep-learning algorithms for object detection.

\subsection{Collaboration \& Gaze Tracking}

Even with the technical challenges in tracking the eye movements of multiple individuals simultaneously, gaze analysis has become a significant component in understanding the interactive and social dynamics in collaborative learning using micro-gaze movements. Joint and mutual attention are important components of collaboration by aligning information acquisition and team dialog \cite{Vatral2022UsingEnvironments}. Social network analysis provides another methodology for measuring team performance by analyzing gaze behaviors \cite{Jing2021EyeCollaboration}. 

In individualistic learning environments, eye-tracking systems typically interface with digital platforms such as computers or mobile devices, where AOIs are defined by on-screen elements \cite{Rajendran2018_Eye-gaze_Learning_Gains}. The digital nature of these AOIs simplifies tracking, as the geometry of on-screen elements can be determined directly from the rendering data, circumventing the need for advanced artificial intelligence (AI) or computer vision techniques. Conversely, collaborative environments present a contrast, often featuring co-located teams interacting with both physical and digital elements. In such settings, tracking of physical AOIs requires sophisticated computer vision algorithms due to the absence of inherent digital geometries to reference.



\begin{figure*}[ht]
    \centering
    \includegraphics[width=\textwidth]{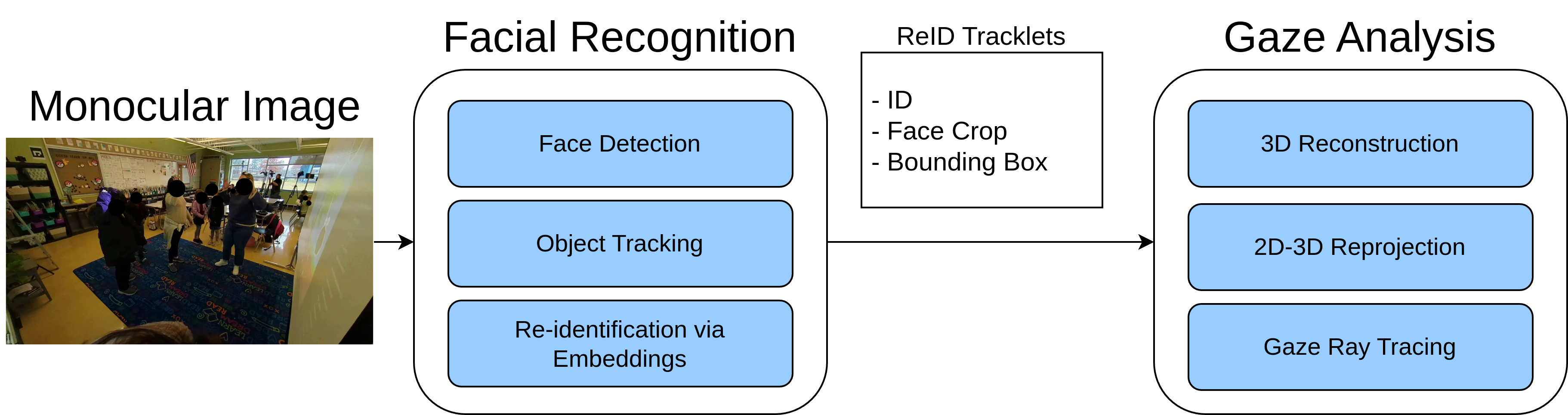}
    \caption{Framework Overview: composed of facial recognition and gaze analysis modules.}
    \Description[Flowchart explaining the input, process, and output of the overall framework]{}
    \label{fig:gaze_3d_analysis_framework}
\end{figure*}

Using computer vision methods for 3D gaze and gaze target estimation marks a progression in collaborative gaze analysis by enabling the capture of multiple gaze data through a single device. However, this approach has its limitations. Traditional 3D gaze estimation techniques primarily provide the 3D rotation, \( R \in \mathbb{R} \), of gaze orientation but lack the 3D translation, \( t \). Moreover, these methods often struggle with targets that move out of frame and face challenges related to video consistency and accurate depth estimation. To overcome these issues, we adopt a method involving 3D reconstruction to maintain spatial consistency across our gaze analysis. This approach ensures that the gaze analysis remains accurate and consistent. Without this additional 3D contextual information, such as the precise locations of the gaze vectors and the environmental layout, the 3D gaze data would typically be projected back onto the 2D image plane, leading to a significant loss of critical spatial information \cite{Guan2023End-to-EndContext, Chong2020DetectingVideo, Tonini2022MultimodalDetection, Tonini2023Object-aware}.

\section{Proposed Method}

Our proposed method is illustrated in Figure \ref{fig:gaze_3d_analysis_framework}. Our framework consists of two main sequential components: (1) a face recognition module and (2) a gaze analysis module. We start with a monocular image and perform face detection employing a finely-tuned Multi-task Cascaded Convolutional Network (MTCNN) model \cite{Zhang2016JointNetworks}. The detected faces are enclosed within bounding boxes, which are then tracked using a custom-designed tracking algorithm. These tracklets serve as input to a re-identification process that utilizes FaceNet \cite{Schroff2015FaceNet:Clustering} to create face embeddings and match them to a pre-established gallery of participant faces.

This automated detection, tracking, and re-identification process allows us to maintain consistent identification across frames, producing reliable bounding boxes for subsequent analysis. In the gaze analysis module, we perform 3D reconstruction by reprojecting the 2D detected data into 3D space. As a next step, we perform 3D gaze ray tracking, which enables us to determine the focus of each participant's gaze and to encode this information in relation to OOIs. This comprehensive approach ensures accurate gaze tracking and analysis by integrating advanced face recognition with sophisticated gaze-tracking techniques, allowing for a deep understanding of participant engagement and focus.

\subsection{Face Recognition Module}

\begin{figure*}
    \centering
    \includegraphics[width=\textwidth]{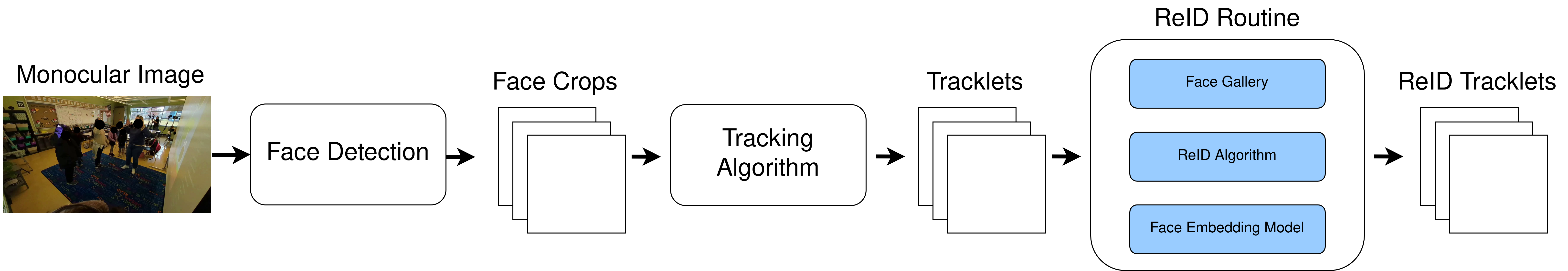}
    \caption{Face Recognition Module: With the consequent use of detecting, tracking, and re-identifying faces, ReID tracklets are generated across a video.}
    \Description[Flowchart showing the processes involved within the face recognition module]{}
    \label{fig:facial_recognition_module}
\end{figure*}

Using off-the-shelf computer-vision models that were fined-tuned to our setting (i.e., a classroom), we track the 2D position of participants (students, teachers, and researchers) in an identify-preserving fashion. The first step of this computational pipeline (see Figure ~\ref{fig:facial_recognition_module}) is the 2D bounding box detection of participants' faces using a fine-tuned MTCNN in a frame-by-frame fashion. To reduce the need to perform a costly re-identification operation, we employed a tracking algorithm that leverages prior knowledge, i.e., bounding boxes of the same object should exhibit small changes from one frame to the next. To implement this, the tracking algorithm uses an Euclidean distance-based approach to match current and prior detections in a computationally efficient manner. Using this tracking strategy we were able to link a large percentage of detections together, except when complete or out-of-frame occlusions caused a failed tracking and a new tracklet ID was generated. For these cases, we had to develop a robust re-identification strategy that used face embeddings and a prior face gallery. 

Our study utilized the DeepFace library and its FaceNet model to create vector embeddings from facial images \cite{serengil2020lightface, Schroff2015FaceNet:Clustering}. We established a reference set of face embeddings for each participant, known as anchor embeddings. When the tracking algorithm lost continuity and initiated a new tracklet ID, the corresponding face crop was embedded and compared against the gallery using cosine similarity, adhering to a stringent identity-matching threshold. Unidentifiable face detections, often due to side profiles or distant captures, were excluded from subsequent analysis to maintain data integrity. The facial recognition module's output is a re-identified (ReID) tracklet that includes participant ID, bounding box coordinates, and the associated face crop, facilitating participant-specific analysis during the inference phase of the computer vision pipeline.

\subsection{Gaze Analysis Module}

\begin{figure*}
    \centering
    \includegraphics[width=\textwidth]{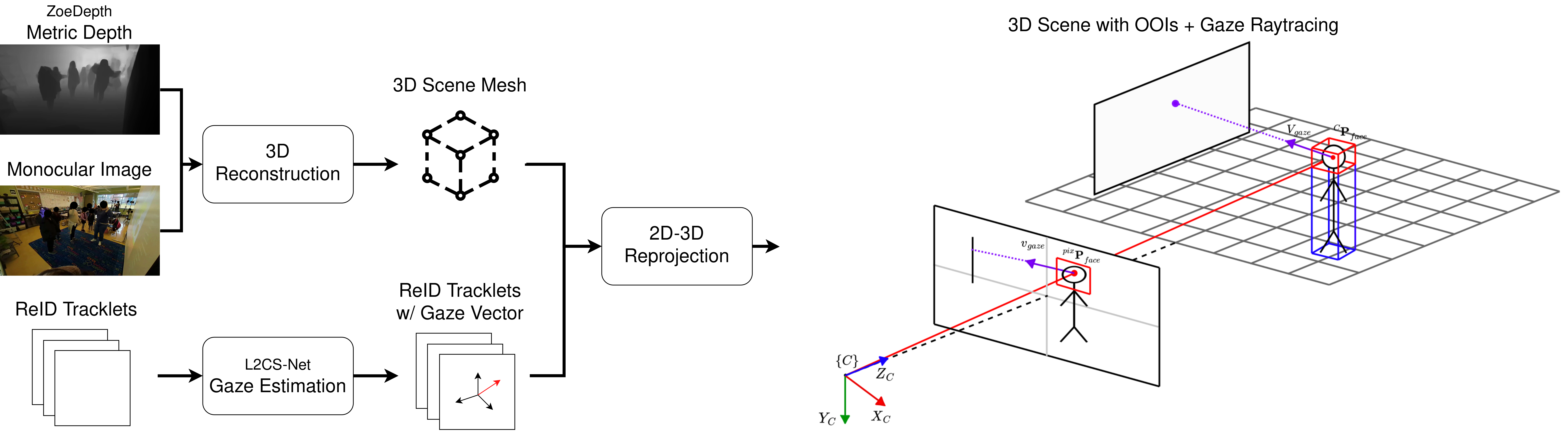}
    \caption{Gaze Analysis Module: Given a monocular image, the depth image is predicted, combined with the ReID tracklets, and then projected into a 3D scene.}
    \Description[Flowchart showing the processes involved within the gaze module]{}
    \label{fig:enter-label}
\end{figure*}

The gaze module utilizes ReID tracklets for each frame to conduct gaze estimation, 3D reconstruction, 2D-3D reprojection, and OOI encoding (see Figure ~\ref{fig:enter-label}). Gaze estimation is achieved using L2CS-Net \cite{Abdelrahman2023L2CS-NetOfficial}, which processes face crops to output 3D gaze vectors represented by pitch and yaw angles. These vectors yield a 3D rotation matrix for gaze orientation, but, by itself, it does not fully capture the 3D gaze information to properly place the gaze within the 3D scene. To overcome this, we employed ZoeDepth \cite{Bhat2023ZoeDepthOfficial}, a metric depth estimation model that provides consistent frame-by-frame depth estimation. By using a metric depth estimation model, we achieve a consistent and properly scaled depth in meters. Unlike other methods that infer depth based on facial or eyeball dimensions, ZoeDepth offers dense pixel-level depth predictions. These predictions facilitate 3D scene reconstruction by creating a mesh from the monocular image, which is updated per frame to accurately represent both dynamic and static elements within the classroom environment.

To effectively reproject our 2D face tracking analyses into 3D gaze tracking, we employ a 2D-3D reprojection method. For each identified feature, such as a bounding box surrounding a face, we project this information into 3D space using a transformation matrix $\mathbf{M}_{face}$. The matrix's rotation component $\mathbf{R}$ is an identity matrix, as rotation adjustment is not required for gaze ray tracing. The translation vector $t$ is determined by the bounding box's centroid ${}^{pix}\mathbf{P}_{face}$ and the depth $z$ is calculated as follows:

\[
t = 
\begin{bmatrix}
    X \\
    Y \\
    Z
\end{bmatrix}
= z \cdot
\begin{bmatrix}
    \frac{x - c_x}{f_x} \\
    \frac{y - c_y}{f_y} \\
    1
\end{bmatrix}
\]

This process results in a 3D bounding box that is specific to the face, and a second bounding box for the full body, which is estimated based on fixed dimensions and orientation. We carefully manage potential mislabeling issues by manually marking the floor plane once per session, especially when participant heights vary. Following this, the gaze vector, originally derived from the L2CS-Net, is transformed into a 3D vector using the established face bounding box as its origin and then converted into a rotation matrix to finalize the 3D placement, denoted as $\mathbf{M}_{gaze}$.

For the 3D scene reconstruction, we manually place static objects such as displays, walls, and floors using a 3D annotation tool, \href{https://github.com/ykzzyk/vision6D}{Vision6D}. With all essential objects placed, we conduct a comprehensive 3D gaze analysis through gaze ray tracking. Utilizing the transformation matrix $\mathbf{M}_{gaze}$, we determine the object of focus by ray tracing the gaze vector and identifying the first object it intercepts, excluding the individual's own face and body. This process is facilitated by Trimesh ray tracing method, encoding each gaze fixation on dynamic participants or static objects within the scene. In more detail, the Trimesh library is a versatile tool that facilitates the loading and processing of triangular meshes, providing utilities for creating, editing, and analyzing 3D geometry in an efficient and straightforward manner. In the context of ray tracing with Trimesh, rays can be cast from a point in space through the mesh, and the library can calculate the points at which the rays intersect the mesh.

The ray tracing mechanism is mathematically represented by:

\[
\mathbf{r}_{\text{transformed}}(t) = \mathbf{M}_{gaze} \begin{bmatrix} \mathbf{o} + t\mathbf{d} \\ 1, \end{bmatrix}
\]

where \(\mathbf{M_{gaze}}\) includes both rotation and translation components. To find the initial point of intersection (\(t_{\text{min}}\)) with any mesh:

\[
t_{\text{min}} = \min \left\{ t \mid t > 0, \mathbf{r}_{\text{transformed}}(t) \cap \text{Mesh} \right\}.
\]

Here, \(\cap\) denotes the intersection between the ray and a mesh, identifying the closest point of contact. This approach enables 3D-consistent tracking of gaze interactions within a complex 3D environment, enhancing our understanding of participant engagement and interaction dynamics.

\section{Results}

To test our approach in an authentic educational setting, we conducted an empirical study in a middle school classroom in the southeastern United States. The students engaged with the X environment, a mixed-reality platform designed to facilitate the enactment of scientific processes by students collaborating in small groups \cite{danish2022designing}. The study was approved by the Institutional Review Board (IRB) at Y University.

Within the X framework, students enact a scientific process, which is displayed as a simulation on a central screen in the classroom. The system is interactive, allowing students to take on roles in the scientific process. Their physical movements in the proximity of the display are translated into interpretable activities in the simulation environment thus helping them interpret their real-world movements in the context of the scientific process being modeled. This interactive design grants students the autonomy to make choices regarding their actions and interactions, not only with the simulation but also with their peers, educators, researchers, and the audience of fellow students.

The dynamic nature of the X environment yields a complex array of interactions encompassing physical movements, gestures, eye contact, emotional expressions, and verbal communication. It is important to collect comprehensive multimodal data to thoroughly understand students' learning behaviors and assess their advances in grasping scientific concepts and processes. This includes video recordings, audio captures, spatial tracking, and detailed logs from the simulation activities conducted within this mixed-reality setting.


\paragraph{Data Collection and Instrumentation}

\begin{figure}
    \centering
    \includegraphics[width=\columnwidth]{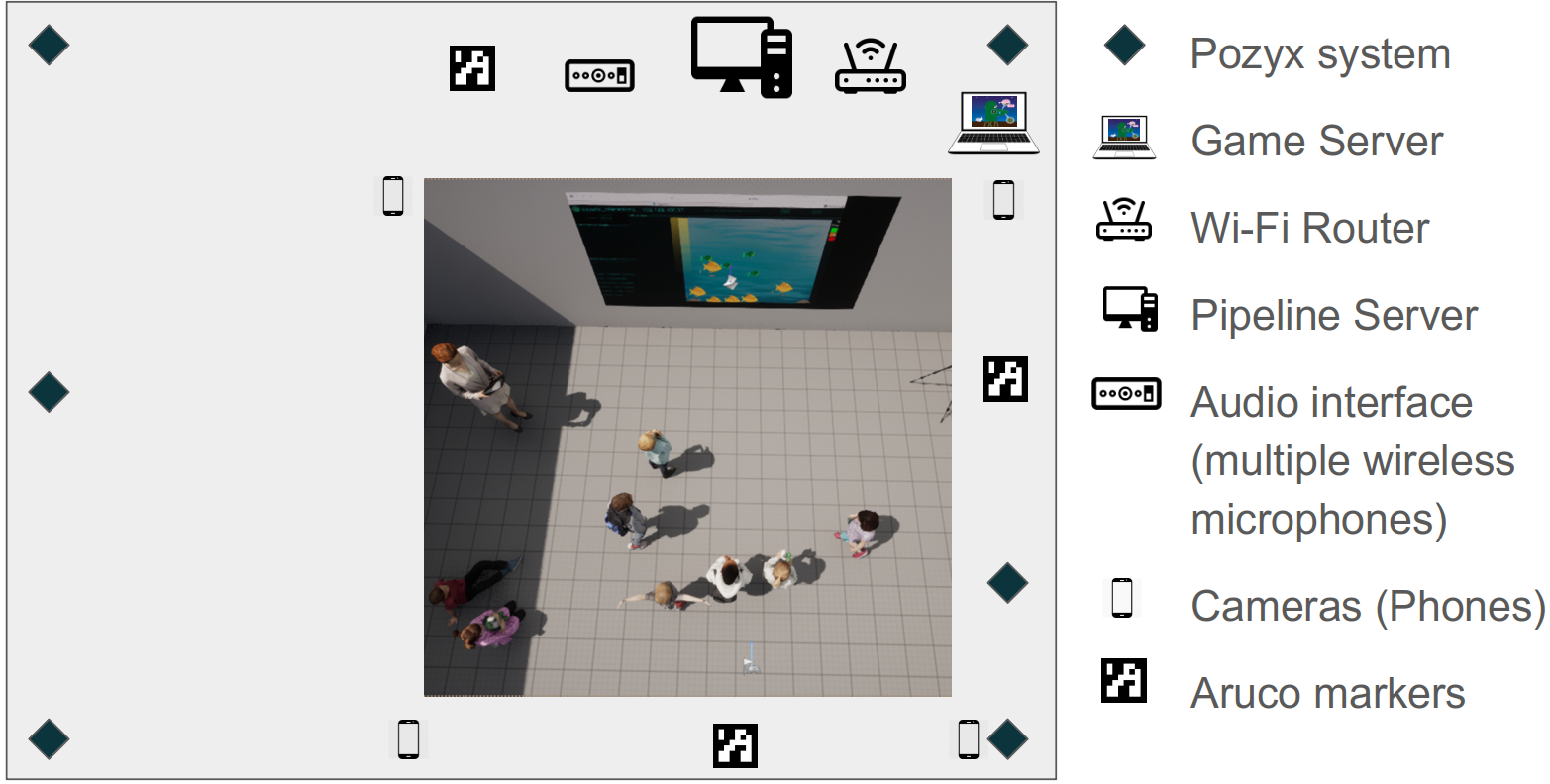}
    \caption{Data Collection and Instrumentation in the X environment.}
    \Description[A top-down layout describing the equipment setup of the X learning environment.]{}
    \label{fig:gemstep_instrumentation}
\end{figure}

\begin{figure*}[t]
    \centering
    \begin{subfigure}[b]{0.8\textwidth}
        \centering
        \includegraphics[width=\textwidth]{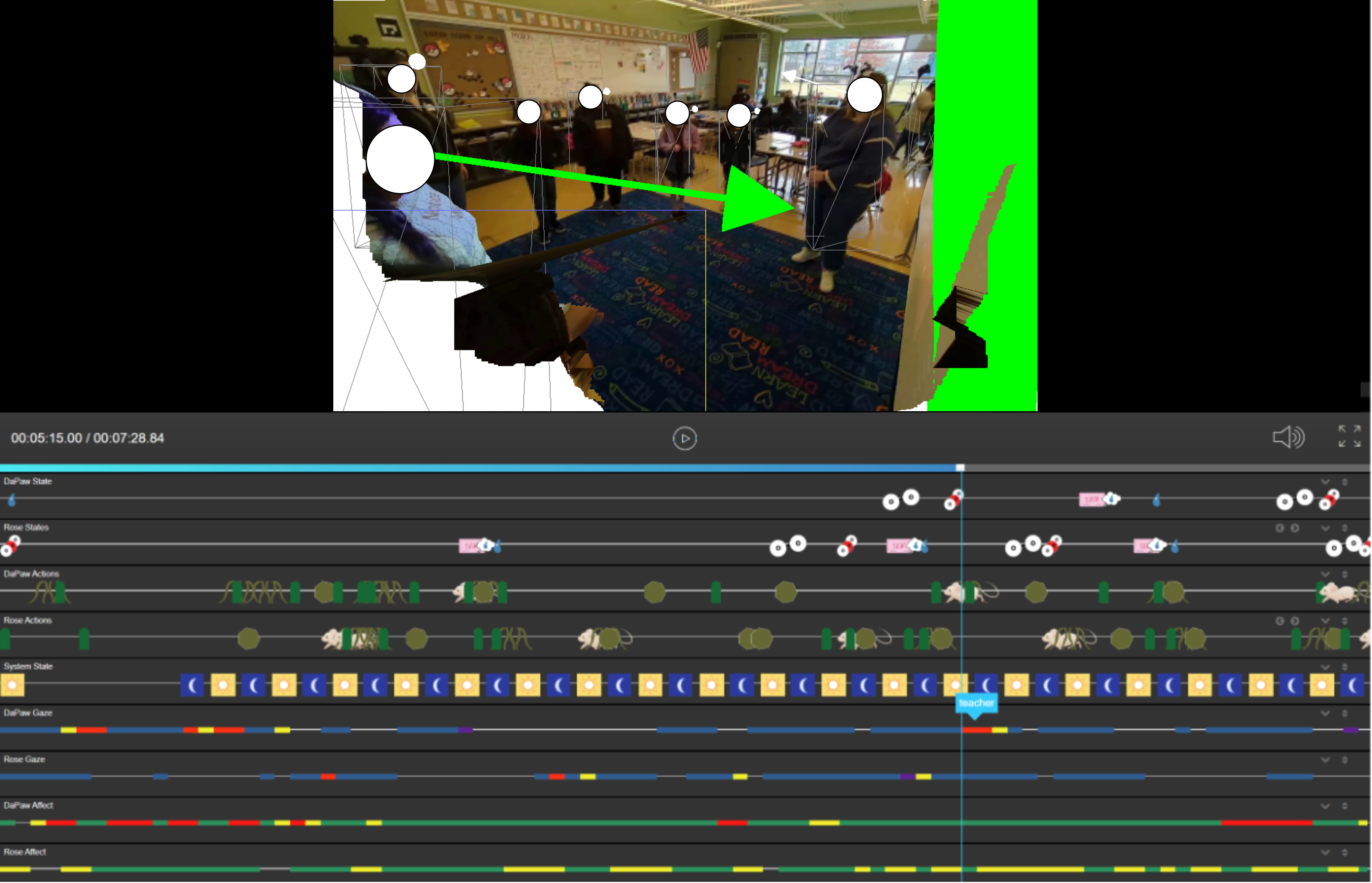}
        \caption{Timeline Example A: Instance of a single student looking at the classroom projector, with the teacher conversing with all students.}
        \label{fig:sub1}
    \end{subfigure}
    \hfill
    \begin{subfigure}[b]{0.8\textwidth}
        \centering
        \includegraphics[width=\textwidth]{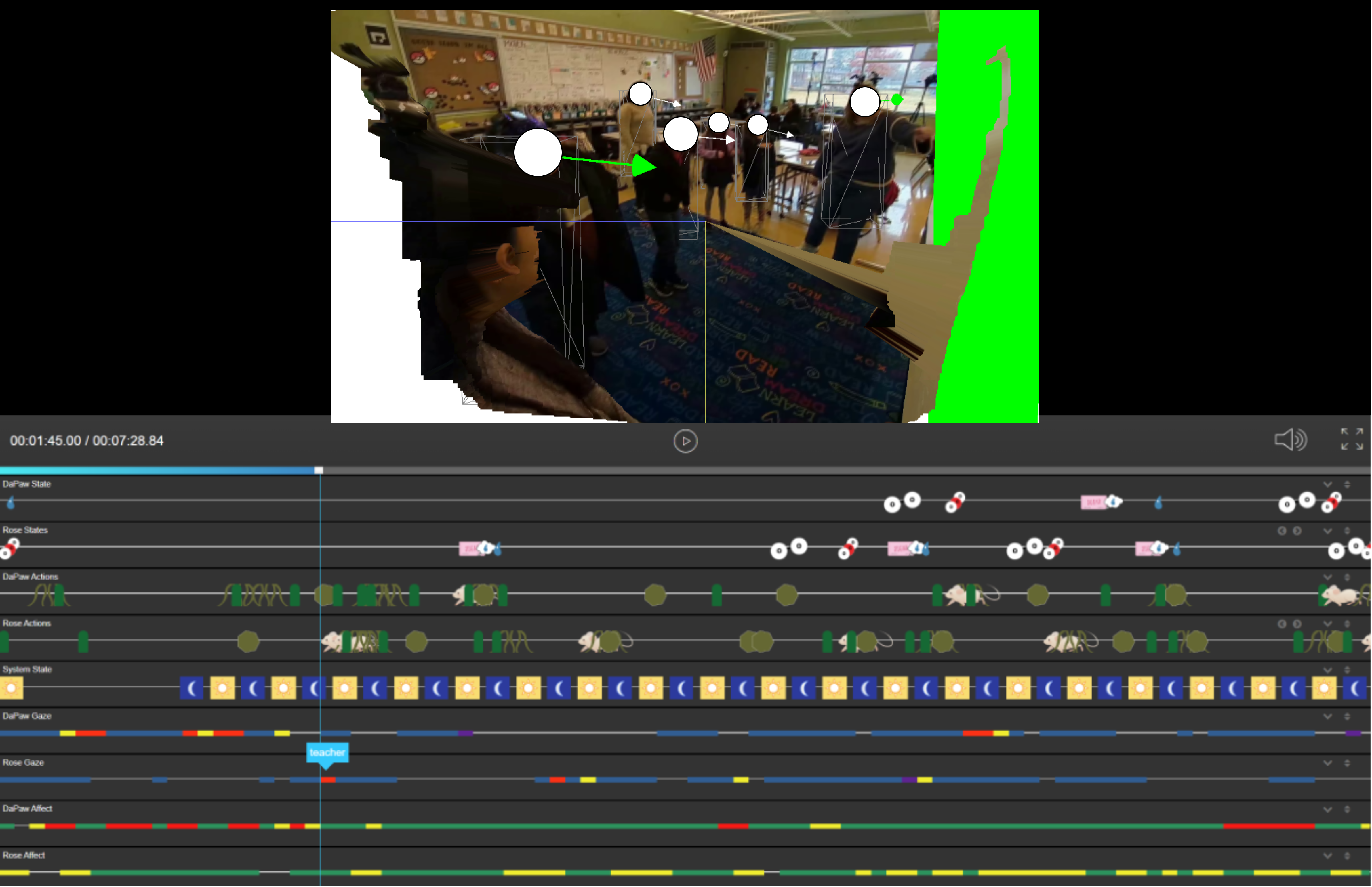}
        \caption{Timeline Example B: Teacher leading instruction by gazing toward the projector and students following her gaze and body language.}
        \label{fig:sub2}
    \end{subfigure}
    \caption{Examples of Multimodal Timelines with gaze visualizations.}
    \Description[Two example images of the timeline describing pedagogical events that were well contextualized using the gaze modality.]{}
    \label{fig:timeline}
\end{figure*}

We employed ChimeraPy \cite{davalos2023chimerapy}, a distributed multimodal data collection framework, to collect data from various devices and data streams. Figure \ref{fig:gemstep_instrumentation} illustrates our equipment setup from a top-down perspective of the room. For data capture, we used Google Pixel 7a mobile phones to record standard monocular videos at a resolution of 1920x1080. Aruco markers were strategically placed within the room to facilitate 3D calibration and ensure consistent positioning across different cameras. Audio data was captured using wireless microphones and subsequently integrated via an audio interface. We also gathered system logs and screen recordings to enable a comprehensive analysis of students' learning behaviors. For this study, we focused on analyzing a single video to assess our methodology, with plans to incorporate a multi-camera setup in future expansions of the research. Our method processed the video frame-by-frame, allowing for detailed gaze analysis of each participant, and maintaining consistent identity tracking throughout the activity.


\paragraph{Recognition Error.} As Figure ~\ref{fig:gaze_3d_analysis_framework} shows, the initial module of the method is facial recognition performed by automatic identify-persistent tracking but errors occurred. Failed re-identification occurs when the top-1 matches face embeddings. 

Our facial recognition module showed a decrease in performance relative to the 99.2\% accuracy benchmark established by FaceNet on the Labeled Faces in the Wild dataset. Notably, the module's accuracy in identifying distinct individuals within our experimental sessions varied. Recognition of the 'TEACHER' participant was commensurate with FaceNet's reported accuracy. Conversely, the accuracy markedly declined to around 80\% for participants `S1' and `S2', who are children. This decrease is likely attributed to the challenges associated with accurately identifying younger facial features, which can differ significantly from adult features that the model is predominantly trained on.

The performance was notably lower for `S3', at just 44\%. This decrease is part of a broader issue related to algorithmic bias in facial recognition technology \cite{khalil2020investigating}. For example, individuals many algorithms show lower accuracy in identifying faces with darker skin complexions. This issue arises because many facial recognition models, including the one we employed, were trained on datasets consisting of mostly white adult faces. This lack of diversity in training data leads to biased performance, where the model fails to generalize effectively across different skin tones and facial structures associated with different ethnicities.

This disparity in recognition accuracy underscores the urgent need for developing facial recognition models that are trained on a more diverse range of datasets. These datasets should encompass a wide variety of ages, skin tones, and facial features to ensure more equitable and effective performance across all user demographics. Such improvements in model training are essential to enhance the inclusivity and fairness of facial recognition technology.


\paragraph{Timeline} We constructed a comprehensive timeline to represent the gaze data throughout the entire video, as depicted at the bottom of Figure \ref{fig:timeline}. This timeline divides the gaze data into 5-second intervals, applying a median pooling strategy with a 2-second threshold to mitigate noise and improve clarity \cite{fontelesAIED2024}. We leveraged our facial recognition system to generate individual timelines for each participant, distinctly illustrating their gaze directed at the teacher, other participants, and objects within the room. During the X game, our analysis showed that participants predominantly focused on the central display of the game, namely the screen. Our system adeptly captured this focal behavior despite considerable movement by the students and teacher, all without the necessity for calibration. Additionally, the timeline exposed gaps primarily resulting from obstructions in the participants' faces, as movements within the environment intermittently blocked the camera's line of sight to certain individuals.

In our study, we employed an interactive multimodal timeline user interface \cite{fontelesAIED2024} to display videos augmented with gaze visualizations in conjunction with the timeline. This approach enables learning sciences researchers to explore the data informatics thoroughly, enabling them to gain a deeper understanding of the learning experiences of both students and teachers. The timeline incorporates data from the X-game state that includes students' gaze and affective states. Our methodology's frame-by-frame analysis yields granular temporal gaze data, facilitating the discernment of overarching trends and behaviors. This level of detail represents an advance over prior studies that primarily utilized heat maps and other aggregate visualization methods for human interpretation and analysis. By presenting the data in a continuous, temporal manner, our interface allows for the swift recognition of changes in patterns and salient events, which can be subsequently scrutinized by human evaluators or artificial intelligence systems.

We have identified two key events in the timeline that illustrate the pedagogical interactions between the students and their teacher. In Figure \ref{fig:sub1}, the teacher is seen engaging multiple students, but one student is focused on the game screen rather than the teacher. While this student may still be listening, the lack of mutual gaze may hinder the student's understanding, especially if the teacher is using body language (e.g., pointing and gestures) or spatial cues to explain science ideas. In another scenario depicted in Fig. \ref{fig:sub2}, the teacher stands by the projector, using hand gestures to highlight aspects of the game and directing attention to the screen with her gaze, indicating its importance in her instructions.

By incorporating body language and speech with gaze data, we capture the nuances of these instructional interactions and their effectiveness. The interactive timeline interface allows researchers and developers to better analyze and develop support systems that enhance teaching and learning experiences through multimodal data integration.

\begin{figure}[t]
    \centering
    \includegraphics[width=\columnwidth]{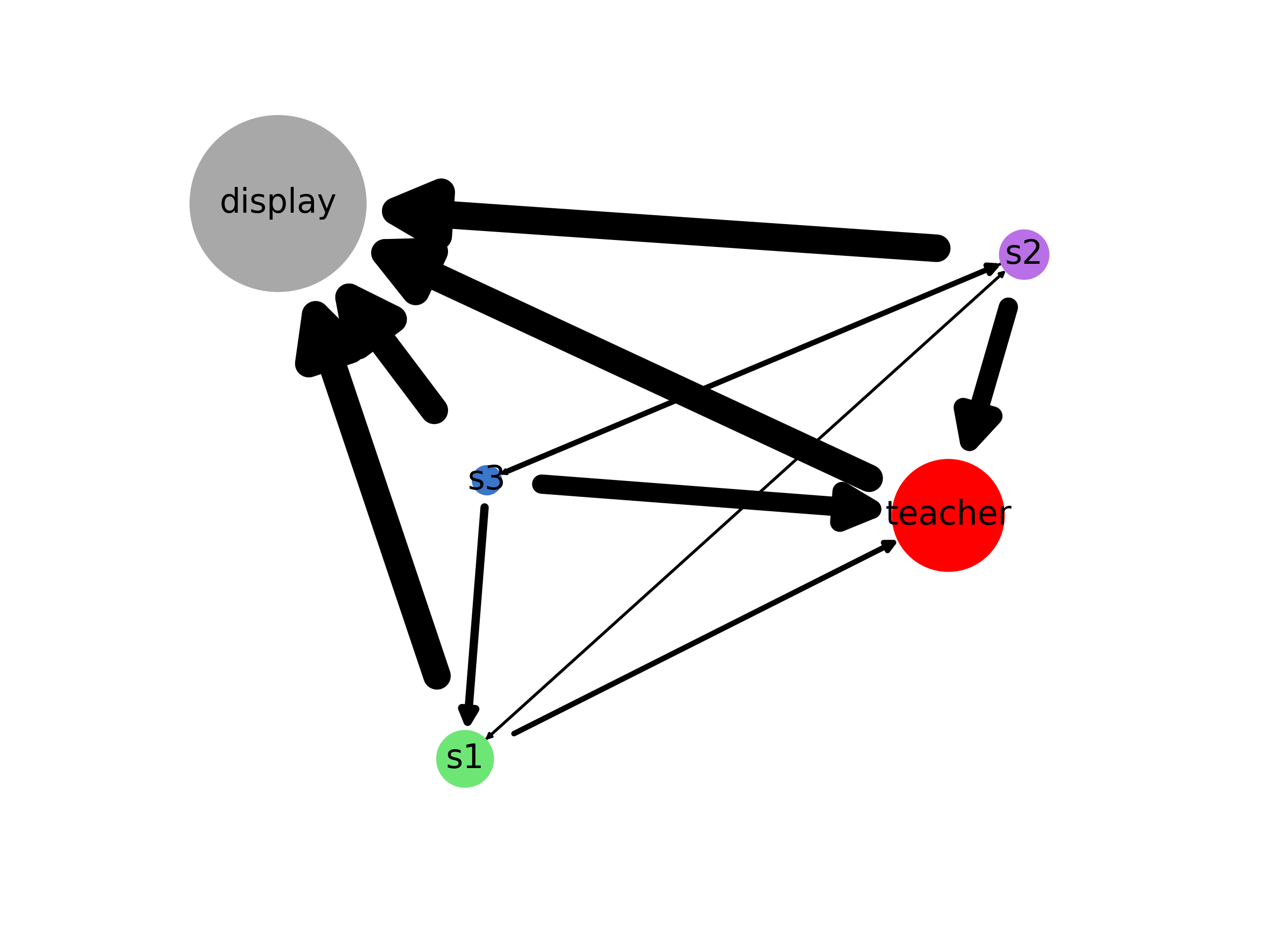}
    \caption{Gaze Attention Network: Describing the observer-observed relationships via a directed graph. The edge weights reflect the total time fixated on the observed, with the node weights equalling the total incoming edge weights.}
    \Description[Network graph based on the observer-observed relations with X game]{}
    \label{fig:gaze_attention_network}
\end{figure}

\paragraph{Network Analysis} By encoding objects of interest (OOIs), tracking participants, and contextually relevant objects, we applied social network analysis using a directed graph. In this graph, the nodes represent the OOIs, sized according to the duration of fixation on each object. The edges illustrate the gaze relationships, with their widths indicating the duration of the observer-observed interaction. Using this method, we constructed an accumulative gaze attention network, depicted in Figure \ref{fig:gaze_attention_network}. Notably, the `DISPLAY' node, central to the game experience, emerged as the most prominent node due to the significant fixation time on this object. While student-to-student gaze was minimal and brief, we observed a noticeable amount of gaze directed from students toward the teacher. Specifically, student `S1' showed the least amount of fixation on both the `TEACHER' and `DISPLAY', suggesting a potential correlation between lower gaze fixation and reduced learning outcomes, which could be explored further to understand the impact on learning processes.

\section{Conclusions}

In this research paper, we present an innovative method that integrates 3D reconstruction with gaze estimation to perform detailed gaze ray tracing. This approach is designed to support scalable OOI encoding, which is a key component of highly contextualized gaze analysis in collaborative settings. Our method leverages advanced computer vision techniques and deep learning models to enhance the traditional gaze estimation processes by incorporating dynamic (such as users) and static (such as props) objects within a three-dimensional space. This allows for a more nuanced understanding of gaze dynamics in mixed-reality environments where both physical and digital elements play critical roles.

The core of our methodology involves a two-step process, starting with a facial recognition module that utilizes a fine-tuned MTCNN model for real-time face detection, an Euclidean-based tracking algorithm, and FaceNet for face re-identification. This is followed by a gaze analysis module that performs a 3D reconstruction of the scene via ZoeDepth and 3D gaze estimation by utilizing L2CS-Net. We employ 3D gaze ray tracing to accurately determine the gaze targets within the environment, thereby enhancing our understanding of interactive dynamics in collaborative tasks.

Our approach offers significant improvements over traditional methods by allowing for the analysis of gaze interactions in a spatially aware context, providing insights into how participants in a study focus on and interact with different elements of their environment. This methodology not only advances the field of gaze analysis but also provides a practical framework for analyzing collaborative interactions in educational and professional settings.

Our empirical studies validate the efficacy of our approach in a real-world classroom environment, showcasing its proficiency in accurately tracking and analyzing gaze patterns among participants during dynamic educational activities. This research paves the way for novel applications of gaze analysis in intricate, real-world settings where deciphering the nexus between attention and interaction is vital.

\paragraph{Limitations} The framework's accuracy suffers from the compounding of errors across multiple machine learning models (detection, tracking, re-identification, gaze estimation, and depth measurement). The accumulation of errors degrades the overall system performance, which is particularly noticeable in facial recognition where misidentification of participants can significantly skew results. To address these inaccuracies, manual corrections of re-identification errors were required, indicating a demand for more refined models that are customized for particular domains.

Depth estimation, while crucial for OOI encoding, introduces additional noise and demands high computational resources, achieving only 3 FPS on an RTX 3090 GPU without techniques to reduce computational complexity. This bottleneck hampers the system's ability to operate in real-time and will mandate high computational resources for real-world deployment.

\paragraph{Future Work} To address these issues, we plan to refine the re-identification process by fine-tuning models specifically for educational domains, using classroom data to enhance the accuracy of participant recognition. Additionally, we aim to incorporate both monocular and stereo cameras to improve the accuracy and reliability of the 3D scene reconstruction. This would potentially allow for computationally efficient and real-time performance by optimizing the depth estimation model/technique. Techniques such as model quantization might be explored to speed up inference, although they typically trade off some accuracy. By implementing these improvements, we hope to minimize the computational load and increase the framework's robustness and applicability in real-time educational settings.

\section{Acknowledgments}

This work has been supported by the National Science Foundation AI Institute Grant No. DRL-2112635.
\bibliographystyle{ACM-Reference-Format}
\bibliography{references, mendeley_ICMI-2024_references}

\end{document}